\pdfoutput=1

\documentclass[11pt]{article}

\usepackage{naacl2021}

\usepackage{times}
\usepackage{latexsym}
\usepackage{subcaption}
\usepackage{amssymb}
\usepackage{graphicx}
\usepackage{comment}
\usepackage{amsmath}
\usepackage{enumitem}

\usepackage{pifont}
\usepackage{multirow}
\usepackage{booktabs}
\usepackage{tablefootnote}
\usepackage[T1]{fontenc}

\usepackage[utf8]{inputenc}

\usepackage{microtype}

\title{Deep Natural Language Processing for LinkedIn Search Systems}

\author{Weiwei Guo, Xiaowei Liu, Sida Wang, Michaeel Kazi,\\ {\bf Zhoutong Fu, Huiji Gao, Jun Jia, Liang Zhang, Bo Long}\\
         \texttt{\{wguo,xwli,sidwang,mikazi,zfu,hgao,jjia,lizhang,blong\}@linkedin.com}\\
         LinkedIn, Mountain View, CA}

\begin{document}
\maketitle
\vspace{-10mm}
\begin{abstract}
Many search systems work with large amounts of natural language data, {\it e.g.}, search queries, user profiles and documents, where deep learning based natural language processing techniques (deep NLP) can be of great help.  In this paper, we introduce a comprehensive study of applying deep NLP techniques to five representative tasks in search engines. Through the model design and experiments of the five tasks, readers can find answers to three important questions: (1) \textit{When is deep NLP helpful/not helpful in search systems?} (2) \textit{How to address latency challenges?} (3) \textit{How to ensure model robustness?}
This work builds on existing efforts of LinkedIn search, and is tested at scale on a commercial search engine. We believe our experiences can provide useful insights for the industry and research communities.
\end{abstract}

\section{Introduction}
\begin{figure}
  \includegraphics[width=75mm]{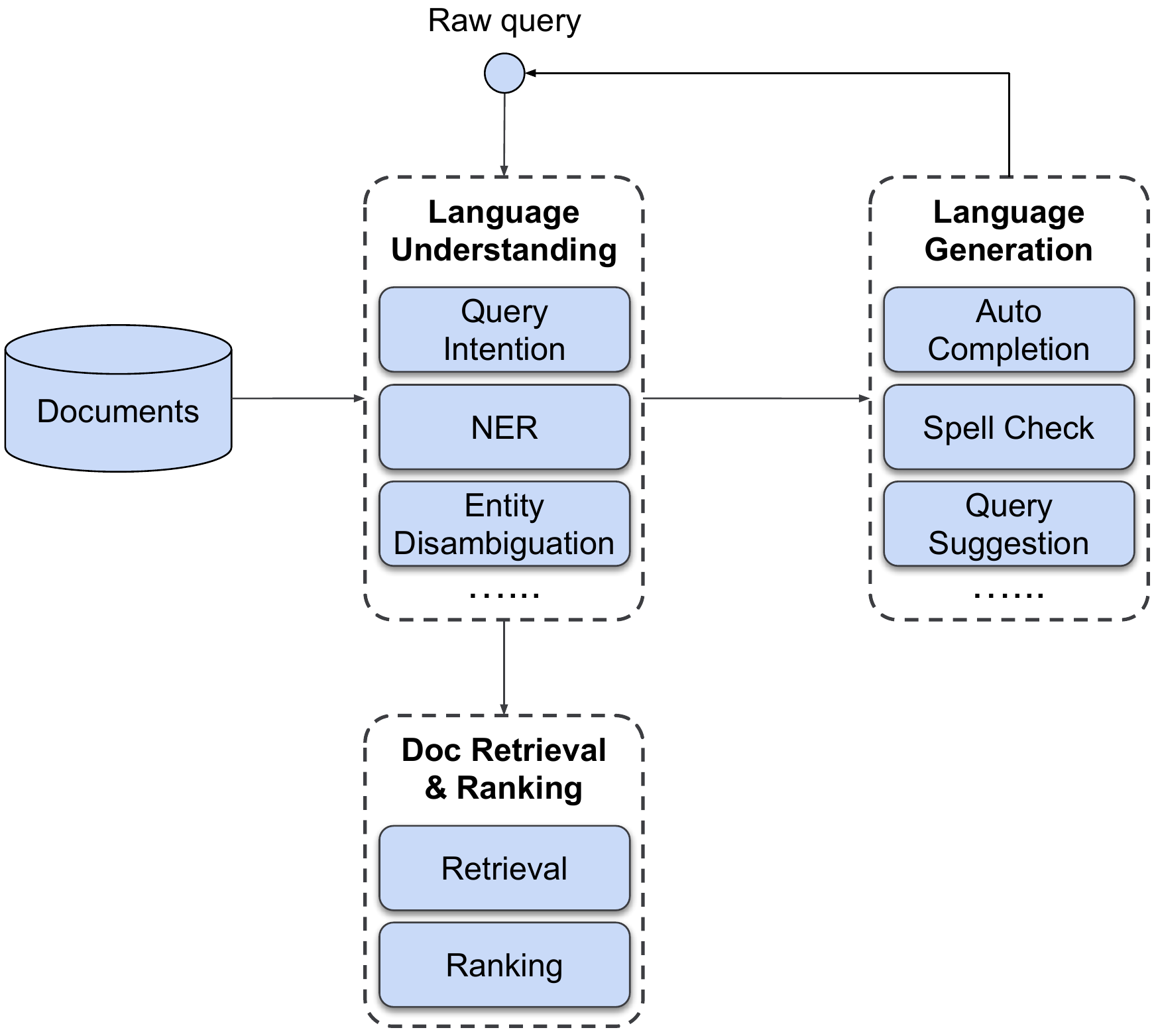}
  \vspace{-2mm}
  \caption{Overview of a search system.}
  \vspace{-2mm}
  \label{figure:overview}
\end{figure}

Search engines are typically complicated ecosystems that contain many components (Figure \ref{figure:overview}), \textit{e.g.}, named entity recognition, query-document matching, etc. A common part of these search components is that they all deal with large amounts of text data, such as queries, user profiles and documents.  

Deep learning has shown great success in NLP tasks \cite{lecun2015}, indicating its potential in search systems. However, developing and deploying deep NLP models for industry search engines requires considering three challenges \cite{Croft2010,mitra2018}. Firstly, serving \textbf{latency} constraints preclude complex models from being deployed in production. In addition, directly trained deep learning models can have \textbf{robustness} issues, which hurts user experience when a model fails on simple queries. The last challenge is \textbf{effectiveness}: often production models are strong baselines that are trained on millions of data examples with many handcrafted features, and have been tweaked for years.

In this paper, we focus on developing practical solutions to tackle the three challenges, and share real world experiences.  As shown in Figure \ref{figure:tasks}, we have picked five representative search tasks that cover the classic NLP problems.  For each task, we investigate the unique challenges, provide practical deep NLP solutions, and share insights from the offline/online experiments. By providing a comprehensive study, we hope the readers can not only learn how to handle the different challenges in search systems, but also generalize and apply deep NLP to new tasks in other industry productions such as recommender systems.

The contribution of the paper is:
\vspace{-2mm}
\begin{itemize}[leftmargin=*]
    \item To our best knowledge, this is the first comprehensive study for applying deep learning in five representative NLP tasks in search engine productions. For each task, we highlight the difference between classic NLP task and search task, provide practical solutions and deploy the models in LinkedIn's commercial search engines.
    \item While previous works focus on offline relevance improvement, we strive to reach a balance between latency, robustness and effectiveness. We summarize the observations and best practice across five tasks into lessons, which would be a valuable resource for search engines and other industry applications.
\vspace{-2mm}
\end{itemize}

\section{Search Systems at LinkedIn}

LinkedIn provides multiple vertical searches, each corresponding to a document type, e.g.,  \textit{people}, \textit{job}, \textit{company}, etc. In this paper, the experiments are conducted on 3 vertical searches (\textit{people search, job search, help center search}), and \textit{federated search}. Federated search retrieves the documents from all vertical searches and blends them into the same search result page.

\begin{figure}
  \includegraphics[width=78mm]{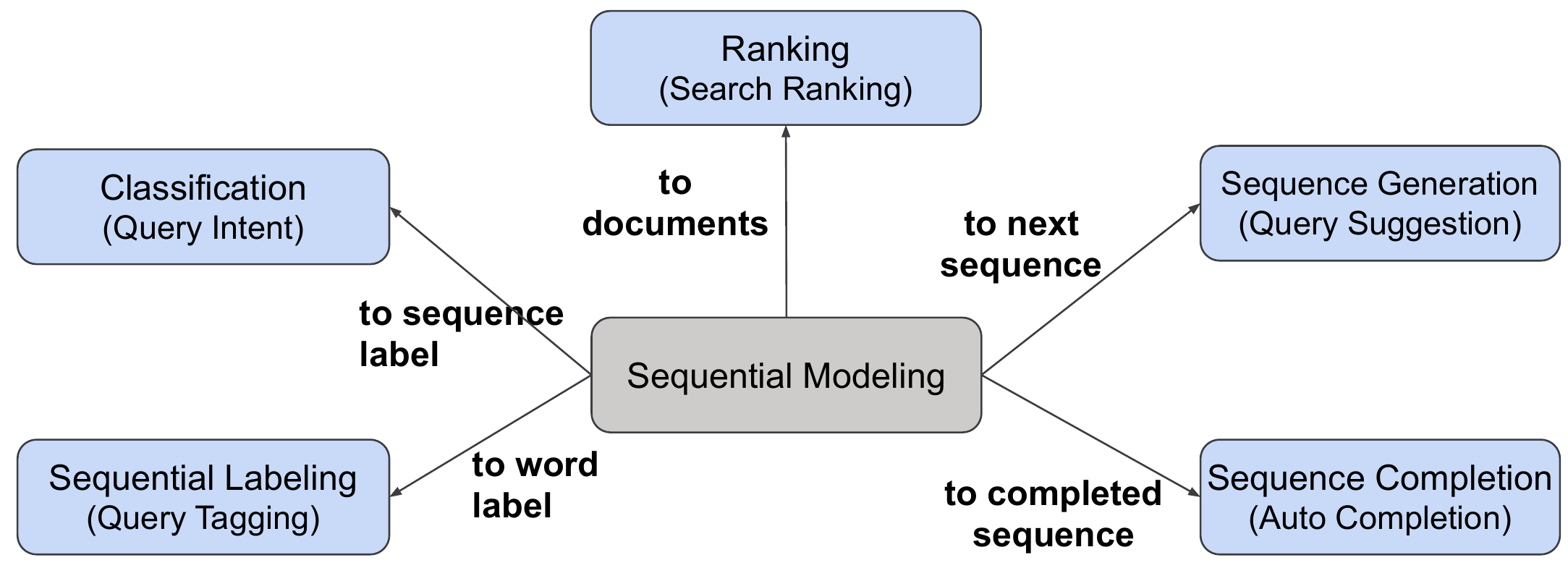}
  \caption{\small Deep NLP models are applied to five representative search tasks.}
  \vspace{-4mm} 
\label{figure:tasks}
\end{figure}

A typical search system is shown in Figure \ref{figure:overview}.  There are three main components: (1) \textit{Language understanding} to construct important features for the other two components; (2) \textit{Language generation} to suggest news queries that are more likely to lead to desirable search results.    (3) \textit{Document retrieval \& ranking} to produce the final results of search systems. Usually the whole process should be finished within several hundreds of milliseconds. In this paper, we selected representative search tasks (Figure \ref{figure:tasks}) that reflect five common NLP tasks: classification, ranking, sequence labeling, sequence completion and generation. 

It is worth noting that search datasets are different from classic NLP task datasets, mainly in two aspects: (1) Data genre. The data unit of classic NLP datasets \cite{pang2005,hu2004} is complete sentences with dozens of words, while queries only have several keywords without grammar. (2) Training data size. Classic NLP datasets are human annotated, therefore the size is usually around tens of thousands of sentences, while search tasks have millions of training examples derived from click-through search log.

\section{Search Tasks}
\label{section:five-task}
In this section, we introduce each specific task, outline the challenges, show how to overcome the challenges, and analyze the offline/online experiment results.
The experiments are conducted on the LinkedIn English market.  Offline results are reported on the test set. All reported online metrics are statistically significant with p<0.05.  

\subsection{Query Intent Prediction}
\noindent\textbf{Introduction: } 
Query intent prediction \cite{kang2003query,hu2009understanding} is an important product in modern search engines. At LinkedIn, query intent is used in federated search, and it predicts the probability of a user's intent towards seven vertical searches: \textit{people}, \textit{job}, \textit{feed}, \textit{company}, \textit{group}, \textit{school}, \textit{event}. 
The predicted intent is an important feature leveraged by downstream tasks such as search result blending \cite{li2008learning} to rank higher the documents from relevant vertical searches.



\noindent\textbf{Approach: } CNN has achieved significant performance gains in text classification problems~\cite{kim2014}.  Therefore, we also use CNN for the query intent task. In our approach, we combine the extracted text embedding with handcrafted features, and use a hidden layer to enable feature non-linearity.



\noindent\textbf{Experiments: }
The label is inferred from click-through behaviors in the search log: if a user clicked on a document from one of the seven verticals, then the query is assigned the corresponding vertical label. We use 24M queries for training, and 50K for dev and test each. We also compare with bidirectional LSTM \cite{graves2013}. For the CNN and LSTM model, the vocabulary size is 100k with word embedding dimension as 64; hidden layer dimension is 200; CNN has 128 filters of window size 3 (tri-grams); LSTM has hidden state size of 128.

\noindent\textbf{Results:}
The baseline model is the production model, logistic regression on bag-of-words features and other handcrafted features. The offline relevance and latency performance are shown in Table \ref{table:offline-qim-cnn}. 
Both CNN and LSTM models outperform the production model, meaning that the features automatically extracted by CNN/LSTM can capture the query intents.  
For online experiments (Table \ref{table:qim-cnn-online}), we choose CNN instead of LSTM, since CNN is faster than LSTM and has similar relevance performance. The online results show CNN increases the job documents click metrics.

\begin{table}
\scriptsize
\centering
\begin{tabular}{lcc}
\toprule
                  & \textbf{Accuracy}      & \textbf{P99 Latency} \\
\midrule
LR (baseline)          & - & -  \\
CNN               & $+1.49\%$  & $+0.45$ms \\
LSTM              & $+1.61\%$  & $+0.96$ms \\
\bottomrule
\end{tabular}
\vspace{-2mm}
\caption{\small Offline experiments of query intent task.}
\vspace{-2mm}
\label{table:offline-qim-cnn}
\end{table}

\begin{table}
\scriptsize
\centering
\begin{tabular}{lc}
\toprule
\textbf{Metrics}          &  \textbf{Percentage Lift} \\
\midrule
CTR@5 of job posts      & $+0.43\%$  \\
\bottomrule
\end{tabular}
\vspace{-2mm}
\caption{\small Online experiments for CNN based query intent over production baseline. CTR@5 calculates the proportion of searches that received a click at top 5 items.
\vspace{-2mm}
\label{table:qim-cnn-online}
}
\end{table}


\subsection{Query Tagging}
\label{section:qt}
\noindent\textbf{Introduction: } The goal of query tagging is to identify the named entities in queries. At LinkedIn, we are interested in 7 types of entities: \textit{first name, last name, company name, school name, geolocation, title, and skill}. After entities are identified, many useful features can be constructed for downstream tasks such as query intent prediction or search ranking.

\noindent\textbf{Approach: }
Query tagging is a named entity recognition task on query data. The production model uses three categories of features: character based, word based and lexicon based. It is worth noting that we are able to extract powerful lexicon features leveraging large amount of user generated data, \textit{i.e.}, collecting the lexicon items from the corresponding fields of 600 million member profiles. Because of this, we choose semi-markov conditional random field (SCRF) \cite{sarawagi2005} to train the production model, which can better exploit lexicon features than CRF \cite{lafferty2001}. 

The bidirectional LSTM-CRF architecture \cite{lample2016} proves to be a successful model on classic NLP datasets \cite{sang2003}.  We further extend it to bidirectional LSTM-SCRF.  Essentially, the deep part, Bidirectional LSTM, is used to replace the word-based features.

\begin{table}
\scriptsize
\centering
\begin{tabular}{lll}
\toprule
 & \textbf{Handcrafted Ftrs} & \textbf{F1}\\
\midrule
SCRF (baseline) & char/word/lexicon & - \\
CRF & char/word/lexicon & $-0.6\%$ \\
SCRF-nolex & char/word & $-6.1\%$ \\
\midrule
LSTM-SCRF & char/lexicon & $-0.3\%$ \\
LSTM-SCRF-all & char/word/lexicon & $-0.1\%$ \\
\bottomrule
\end{tabular}
\vspace{-2mm}
\caption{\small Query tagging results, measured by F1 score.}
\label{table:qt-res}
\vspace{-2mm}
\end{table}

\noindent\textbf{Experiments: }
Queries from LinkedIn federated search are collected and manually annotated: 100K training queries, 5K for dev and test each. For LSTM based models, word embedding size is 50; hidden size is 50.

\noindent\textbf{Results:} The traditional method SCRF achieves the best results in Table \ref{table:qt-res}. LSTM-SCRF has all the handcrafted features except word based features, however, it cannot outperform the SCRF baseline, and neither can LSTM-SCRF-all with all handcrafted features.  Due to no significant offline experiment gain, these models are not deployed online.

We believe the major reason is the strength of lexicon features, therefore LSTM does not help much. The SCRF-nolex (without lexicon based features) performance also indicates lexicon features are the most important. Meanwhile, looking at the data, we found most entities are already covered by the lexicons that are built on large scales of data. Other reasons could be due to the data genre: Queries are much shorter than natural language sentences. Therefore, LSTM's ability to extract long distance dependencies is not helpful in this task. 

\subsection{Query auto completion}
\noindent\textbf{Introduction: }
Query auto completion \cite{bar2011} is a language generation task.  The input is a prefix typed by a user, and the goal is to return a list of completed queries that match the user's intent.
Query auto completion has a strict latency requirement, as the model needs to return results for each keystroke.

\noindent\textbf{Approach: }
The traditional auto completion system has two separate steps: candidate generation and candidate ranking \cite{Cai:16}. In our production model, candidate generation use heuristics to extend a query prefix to a set of completed query candidates  \cite{mitra2015}. Candidate ranking uses xgboost \cite{chen2016} to rank the candidates with handcrafted features, with the most effective feature being the query frequency. 

Since query auto completion is a language generation task, an ideal model is neural language modeling \cite{Mikolov2010} to score the probability of a completed query \cite{park2017}.  While this approach achieves impressive relevance results, the computation time is too long for production deployment \cite{Wang2018}.

To solve the latency challenge, we apply unnormalized language model \cite{sethy2015} to candidate ranking phase. This is motivated by the observation that majority of time is spent on the probability normalization over all vocabulary. The normalization factor is approximated by:

\vspace{-3mm}
{\scriptsize
\begin{align*}
\log{P(w_i|h_i)}&=\log{\frac{\exp(v_i^{\top}h_i)}{\sum_j{\exp(v_j^{\top}h_i)}}}\\
&=v_i^{\top}h_i-\log{\sum_j{\exp(v_j^{\top}h_i)}}=v_i^{\top}h_i-b
\end{align*}
}%
, where $b$ is another parameter to estimate.


\begin{table}
\scriptsize
\centering
  \begin{tabular}{lcc}
    \toprule
    \textbf{Ranking Models} & \textbf{MRR@10} & \textbf{P99 Latency}\\
    \midrule
    xgboost  & - & - \\
    Language model & $+3.2\%$ & +53ms \\
    Unnormalized LM & $+3.1\%$ & +3ms \\
    \bottomrule
  \end{tabular}
  \vspace{-2mm}
  \caption{\small Offline experiments of query auto completion task. MRR@10 is the mean reciprocal rank of the correct answer in top 10 list. P99 latency is the 99 percentile latency.}
  \vspace{-2mm}
  \label{table:qac-offline}
\end{table}

\begin{table}
\scriptsize
\centering
  \begin{tabular}{lc}
    \toprule
    \textbf{Metric} & \textbf{Lift}\\
    \midrule
    Job Views & +0.43\% \\
    Job Applies & +1.45\% \\
  \bottomrule
  \end{tabular}
  \vspace{-2mm}
\caption{\small Online experiments of query auto completion on job search. "Job Views"/"Job Applies" is the number of job posts clicked/applied from search, respectively.}
\vspace{-2mm}
\label{table:qac-online}
\end{table}

\noindent\textbf{Experiments}: The model is applied on job search query auto completion.  We follow the experiment setting and the training/dev/test splitting schema defined in \cite{mitra2015}.
The LSTM has one layer with a 100 dimension hidden state. The vocabulary is 100K words with 100 dimensional embeddings.

Table \ref{table:qac-offline} and \ref{table:qac-online} show the offline/online results, respectively. Compared to language model approach, our unnormalized language model approach can reach the same level of relevance performance,  while significantly reducing the latency. 

\subsection{Query Suggestion}
\noindent\textbf{Introduction: }
Query suggestion \cite{fonseca2005,cao2008} is another essential part of our search experience. Many search engines offer such a function, \textit{e.g.},  Google's "Searches related to ...", and LinkedIn's "People also search for", to assist users to seek relevant information.  

\noindent\textbf{Approach: }
Our production approach is based on frequency counting of search queries. To be specific, it collects query reformulation pairs from search log, and for each input query, sorts the suggestions by the query pair frequency. Heuristics are used to make sure the query pairs are semantically related: (1) the query pairs must be in the same session, where sessions are defined by queries separated by no more than 10 minutes; (2) two queries must share one common word.

We formulate the problem as machine translation in the sequence-to-sequence (seq2seq) framework \cite{Sutskever2014}, of which the main benefit is to generalize to infrequent and unseen queries. We find the deep learning model can overfit to  reformulation pairs if we are not careful.  When the training data contains a query \textit{generalization}, e.g. "research scientist --> scientist", the trained model will degrade to only delete words, to produce low perplexity suggestions. We handle this by removing these types of examples from the training data.

The seq2seq model has a large latency, which can be an issue in production. In this case, we serve our model in parallel with search result ranking, which gives us plenty of time (more than 100ms) to run the seq2seq model.

\noindent\textbf{Experiments: }
The seq2seq model is tested in federated search. Our baseline is a frequency based method. Training/dev/test data size is 300M/50K/50K query pairs. For the seq2seq model, we use a small model to reduce latency (100 hidden size and 2 layers in LSTM).

Our offline experiments in Table \ref{table:qs-offline} demonstrate that the seq2seq can find better candidates, measured by MRR@10. The coverage of the deep learning method is trivially 100\%. The online experiments (Table \ref{table:qs-online}) display evidence of an enhanced user experience, particularly on finding jobs where seq2seq can generate in-domain terminology for novice users.

\begin{table}
\scriptsize
\centering
\begin{tabular}{lcc}
\toprule
 & \textbf{MRR@10} & \textbf{Coverage} \\
\midrule
Frequency baseline & - & 67.3\% \\
Seq2Seq & +11.1\% & 100\% \\
\bottomrule
\end{tabular}
\vspace{-2mm}
\caption{\small Offline experiments for query suggestion task.}
\vspace{-2mm}
\label{table:qs-offline}
\end{table}

\begin{table}
\scriptsize
\centering
\begin{tabular}{lc}
\toprule
\textbf{Metric} & \textbf{Lift}  \\
\midrule
Job Views & $+0.6\%$ \\
Job Applies & $+1.0\%$ \\
\bottomrule
\end{tabular}
\vspace{-2mm}
\caption{\small Online performance of query suggestion task.}
\vspace{-2mm}
\label{table:qs-online}
\end{table}

\subsection{Document Ranking}
\noindent\textbf{Introduction: } Given a query, a searcher profile and a set of retrieved documents, the goal of document ranking is to assign a relevance score to each document and generate the ranking. 
Latency is the biggest challenge for this task, since the data unit of document ranking task is thousands of documents. The other challenge comes from effectiveness.  The production model is optimized in many iterations, with many strong handcrafted features.

We conduct experiments on people search and help center. People search has 600M member profiles, and help center has 2700 FAQ documents.

\noindent\textbf{Approach: }
The production model uses xgboost \cite{chen2016} for training. In people search, many non-text based features are used, \textit{e.g.}, personalized features based on social network structures, past user behaviors, document popularity, etc.

\begin{figure}
  \includegraphics[width=0.35\textwidth]{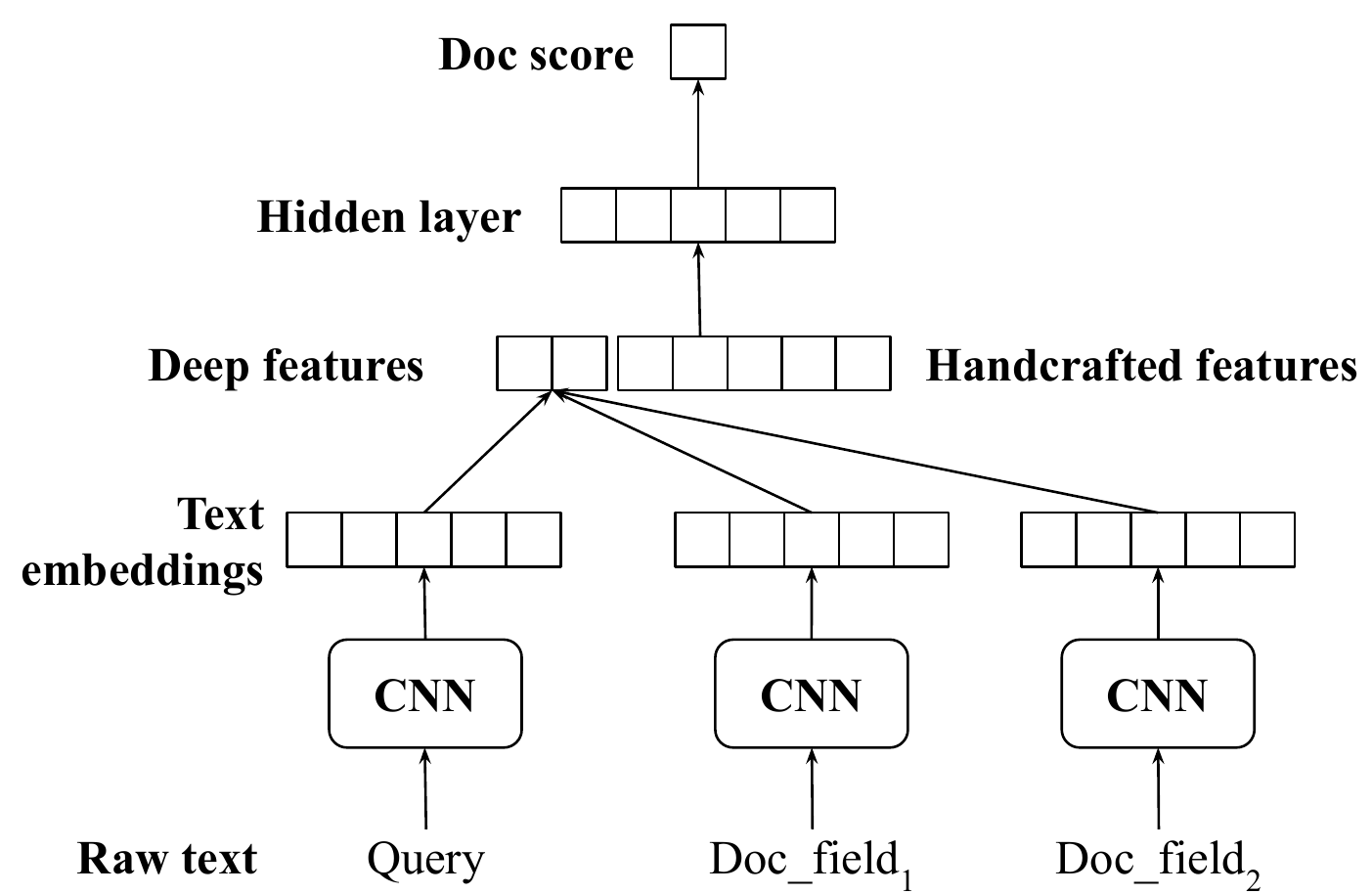}
  \caption{\small Model architecture for the ranking model. The learning-to-rank layer is not drawn in the figure.}
  \label{figure:ranking-model}
\end{figure}

One benefit of the neural networks approaches is that we can easily combine the deep NLP based semantic matching with other techniques that prove to be effective. The architecture is shown in Figure \ref{figure:ranking-model}. To improve the relevance performance, we extend the previous work \cite{Huang2013} by (1) better representing the document by multiple fields; (2) combining hand crafted features with deep features to outperform the production baseline; (3) adding a hidden layer to extract non-linear features.

\noindent\textbf{Online Production Deployment: }
As mentioned before, latency is a major issue for document ranking tasks. We design practical online deployment strategies to reduce computation time. For the help center task, since there are only 2,700 documents, we pre-compute the document embeddings. For people search with 600M member profiles, document pre-computing requires nontrivial infrastructure changes, \textit{i.e.}, a lot of space to store the embeddings and complicated designs to keep them fresh. Our solution is a two-pass ranking strategy: firstly apply a lightweight model without deep learning modules, and choose the top hundreds of documents to send to deep models for reranking. The latency impact over xgboost is shown in Table \ref{table:ranking-latency}.

\begin{table}
\centering
\scriptsize
  \begin{tabular}{lcc}
    \toprule
    \textbf{Search} & \textbf{Deployment Strategy} & \textbf{P99 latency}\\
    \midrule
    People search & Two pass ranking & +21ms\\
    & All-decoding & +55ms \\
    \midrule
    Help center & Document pre-computing & +25ms\\
  \bottomrule
\end{tabular}
\vspace{-2mm}
\caption{\small{P99 latency over xgboost in document ranking.}}
\label{table:ranking-latency}
\vspace{-2mm}
\end{table}

\noindent\textbf{Experiments: }  Training data size is 5M/340K queries for people search/help center, respectively, with 50K for dev and test for both tasks. On average each query has around 10 documents. For all models that apply, batch size is 256; word embedding size is 64.

\begin{table}
\tiny
\centering
  \begin{tabular}{lccc}
    \toprule
     \textbf{Models} & \textbf{People Search} & \textbf{Help Center}\\
    \midrule
    xgboost (baseline) & - & - \\
    CNN-ranking &  $+3.02\%$ & $+11.56\%$\\
    CNN-ranking w/o handcrafted ftrs &  $-4.52\%$ & $+11.07\%$\\
  \bottomrule
\end{tabular}
\vspace{-2mm}
\caption{\small Offline experiments for document ranking task, measured by NDCG@10.}
\vspace{-2mm}
\label{table:ranking-offline}
\end{table}

\begin{table}
\scriptsize
\centering
\begin{tabular}{llc}
\toprule
\textbf{Search} & \textbf{Metrics} & \textbf{Percentage lift} \\
\midrule
\multirow{2}{*}{People Search} & CTR@5 & $+1.13\%$\\
 & Sat Click & $+0.76\%$\\
\midrule
Help Center & Happy Path Rate & $+15.0\%$\\
\bottomrule      
\end{tabular}
\vspace{-2mm}
\caption{\scriptsize Online experiments for document ranking task. "Sat click" is the number of satisfactory searches in people search, such as connecting, messaging or following a profile. "Happy path rate" is a help center search metric: proportion of users who searched and clicked a document without using help center search again in that day.}
\vspace{-2mm}
\label{table:ranking-online}
\end{table}

We report offline and online results in Table \ref{table:ranking-offline} and \ref{table:ranking-online}, respectively. In summary, the offline and online performance are consistent. 
Our model \textit{CNN-ranking} significantly outperforms the baseline by a large margin. By comparing CNN-ranking vs baseline across the two searches, it is interesting to see that the improvement of deep NLP models is significantly larger in the help center setting.  This is caused by the data genre. In help center, the queries and documents are mostly natural language sentences, such as a query "how to hide my profile updates" to document "Sharing Profile Changes with Your Network", where CNN can capture the semantic meaning. In people search, it is more important to perform exact matching, for example, query "facebook" should not return member profiles who work at "linkedin".



\section{Lessons Learned}

In this section, we go beyond the task boundary and generalize the learnings into lessons.  We believe these lessons are not limited to search systems, but also useful for other industry applications such as recommender systems.

\subsection{When is Deep NLP Helpful?}
In general, the deep NLP models achieve better relevance performance than traditional methods in search tasks, and are particularly powerful in the following scenarios:\\
(1) \textbf{Language generation tasks}. Based on the offline experiments, query suggestion (Table \ref{table:qs-offline}) benefits most from deep NLP modeling. This is determined by the nature of language generation tasks. 
In the seq2seq framework, the hidden state summarizes the context well, and the decoder can produce related queries for any input query.\\ 
(2) \textbf{Data with rich paraphrasing}.  In document ranking, the improvement brought by using CNN is larger in help center than in people search (Table \ref{table:ranking-offline} and \ref{table:ranking-online}), since help center has a lot of natural language queries while people search queries are combination of keywords.

\subsection{When is Deep NLP not Helpful?}
No gain is observed when applying LSTM-SCRF for query tagging task. It reveals an important difference between industry and academic data -- huge amount of user generated data. Hence, the SCRF approach can do well with a lexicon built on 600M member profiles. Another factor is data genre. Compared to classic NER datasets \cite{sang2003} with sentence level data points, search queries not have much long distance dependency for LSTM to extract entities.

\subsection{Latency is the Biggest Challenge}
We believe latency to be the biggest challenge of applying deep NLP models to search productions. In this paper, we summarize as below the practical solutions to reduce latency:\\
(1) \textbf{Algorithm redesign}. In query auto completion, we apply the unnormalized language model, which significantly reduces computation time and maintains the same level of relevance.\\
(2) \textbf{Parallel computing}.  In query suggestion, the seq2seq model runs in parallel with search ranking, which leave more than 100ms buffer for seq2seq decoding.\\
(3) \textbf{Embedding pre-computing}. In help center search ranking, the document embeddings are pre-computed, leaving only query processing to be computed at run-time.\\
(4) \textbf{Two pass ranking}.  In people search ranking, a lightweight model is applied to handle all documents, then a deep model reranks the top ones.  Compared to embedding pre-computing, two pass ranking has less of an infrastructure requirement.

\subsection{How to Ensure Robustness?}
Robustness is an essential requirement for industry productions, as it directly affects user experience. Compared to traditional methods, deep NLP models are more likely to overfit the training data and over generalize word semantics.  Our solutions are:
(1) \textbf{Training data manipulation}. In query suggestion, we observe that with generalization pairs ("senior research scientist" -> "research scientist"), the trained seq2seq will mostly generate queries by deleting original words.  We address it by removing the generalization pairs in the training data.\\
(2) \textbf{Reuse handcrafted features}. In document ranking, the deep NLP model may rank the topically related documents higher than the keywords matched documents. For example, in people search, matching professor profiles to query "student". This can be alleviated by incorporating the existing handcrafted features (\textit{e.g.}, keywords matching features such as cosine similarity) into neural networks.

\section{Related Work}
Shallow models are popular in today's search engines. The models are easy to deploy because of small latency and robustness, and they also achieve desirable relevance performance due to the large scale of user generated training data. Recently, deep NLP models have been designed for many of the search tasks \cite{hashemi2016,He2016,Huang2013,Park:17}. These works concentrate on improving offline relevance performance by incorporating additional information such as personalization \cite{Jaech2018,fiorini2018}, session \cite{Dehghani:2017,Ren2018}, etc. There are some existing efforts that are deployed to online productions, however, most models are for the document ranking task with embedding pre-computing approach \cite{ramanath2018,yin2016,Grbovic2018,li2019}. In contrast, we target at productionizing deep NLP models for a set of much boarder tasks by reaching a balance among latency, robustness, effectiveness.

\section{Conclusions}
Industry provides its own set of challenges for deep NLP, different in key ways from classic studied NLP tasks: (1) the amount and type of data, (2) constraints on latency or infrastructure, and (3) the highly optimized production baselines. This paper focuses on how to apply deep NLP models to five representative search productions, illuminating the  challenges along the way.  All resulting models except query tagging are deployed in LinkedIn's search engines.   
More importantly, we also summarize the lessons learned across the tasks.  We listed the factors to consider when estimating the potential improvement brought by deep NLP models for a new task. We show robustness and overfitting can be typically handled with careful data analysis. Latency, which is almost always a concern, can be addressed creatively in many ways, such as algorithm simplification, two-pass systems, embedding pre-computing or parallel computation. 

\bibliographystyle{acl_natbib}
\bibliography{deepnlp}

\begin{thebibliography}{37}
\expandafter\ifx\csname natexlab\endcsname\relax\def\natexlab#1{#1}\fi

\bibitem[{Bar-Yossef and Kraus(2011)}]{bar2011}
Ziv Bar-Yossef and Naama Kraus. 2011.
\newblock Context-sensitive query auto-completion.
\newblock In \emph{WWW}.

\bibitem[{Cai and De~Rijke(2016)}]{Cai:16}
Fei Cai and Maarten De~Rijke. 2016.
\newblock A survey of query auto completion in information retrieval.
\newblock \emph{Foundations and Trends{\textregistered} in Information
  Retrieval}.

\bibitem[{Cao et~al.(2008)Cao, Jiang, Pei, He, Liao, Chen, and Li}]{cao2008}
Huanhuan Cao, Daxin Jiang, Jian Pei, Qi~He, Zhen Liao, Enhong Chen, and Hang
  Li. 2008.
\newblock Context-aware query suggestion by mining click-through and session
  data.
\newblock In \emph{KDD}.

\bibitem[{Chen and Guestrin(2016)}]{chen2016}
Tianqi Chen and Carlos Guestrin. 2016.
\newblock Xgboost: A scalable tree boosting system.
\newblock In \emph{KDD}.

\bibitem[{Croft et~al.(2010)Croft, Metzler, and Strohman}]{Croft2010}
W.~Bruce Croft, Donald Metzler, and Trevor Strohman. 2010.
\newblock \emph{Search engines: Information retrieval in practice}.
\newblock Reading: Addison-Wesley.

\bibitem[{Dehghani et~al.(2017)Dehghani, Rothe, Alfonseca, and
  Fleury}]{Dehghani:2017}
Mostafa Dehghani, Sascha Rothe, Enrique Alfonseca, and Pascal Fleury. 2017.
\newblock Learning to attend, copy, and generate for session-based query
  suggestion.
\newblock In \emph{CIKM}.

\bibitem[{Fiorini and Lu(2018)}]{fiorini2018}
Nicolas Fiorini and Zhiyong Lu. 2018.
\newblock Personalized neural language models for real-world query auto
  completion.
\newblock In \emph{NAACL}.

\bibitem[{Fonseca et~al.(2005)Fonseca, Golgher, P{\^o}ssas, Ribeiro-Neto, and
  Ziviani}]{fonseca2005}
Bruno~M Fonseca, Paulo Golgher, Bruno P{\^o}ssas, Berthier Ribeiro-Neto, and
  Nivio Ziviani. 2005.
\newblock Concept-based interactive query expansion.
\newblock In \emph{CIKM}.

\bibitem[{Graves et~al.(2013)Graves, Jaitly, and Mohamed}]{graves2013}
Alex Graves, Navdeep Jaitly, and Abdel-rahman Mohamed. 2013.
\newblock Hybrid speech recognition with deep bidirectional lstm.
\newblock In \emph{IEEE workshop on automatic speech recognition and
  understanding}.

\bibitem[{Grbovic and Cheng(2018)}]{Grbovic2018}
Mihajlo Grbovic and Haibin Cheng. 2018.
\newblock Real-time personalization using embeddings for search ranking at
  airbnb.
\newblock In \emph{KDD}.

\bibitem[{Hashemi et~al.(2016)Hashemi, Asiaee, and Kraft}]{hashemi2016}
Homa~B. Hashemi, Amir Asiaee, and Reiner Kraft. 2016.
\newblock Query intent detection using convolutional neural networks.
\newblock In \emph{WSDM, Workshop on Query Understanding}.

\bibitem[{He et~al.(2016)He, Tang, Ouyang, Kang, Yin, , and Chang}]{He2016}
Yunlong He, Jiliang Tang, Hua Ouyang, Changsung Kang, Dawei Yin, , and
  Yi~Chang. 2016.
\newblock Learning to rewrite queries.
\newblock In \emph{CIKM}.

\bibitem[{Hu et~al.(2009)Hu, Wang, Lochovsky, Sun, and
  Chen}]{hu2009understanding}
Jian Hu, Gang Wang, Fred Lochovsky, Jian-tao Sun, and Zheng Chen. 2009.
\newblock Understanding user's query intent with wikipedia.
\newblock In \emph{WWW}.

\bibitem[{Hu and Liu(2004)}]{hu2004}
Minqing Hu and Bing Liu. 2004.
\newblock Mining and summarizing customer reviews.
\newblock In \emph{KDD}.

\bibitem[{Huang et~al.(2013)Huang, He, Gao, Deng, Acero, and Heck}]{Huang2013}
Po-Sen Huang, Xiaodong He, Jianfeng Gao, Li~Deng, Alex Acero, and Larry Heck.
  2013.
\newblock Learning deep structured semantic models for web search using
  clickthrough data.
\newblock In \emph{CIKM}.

\bibitem[{Jaech and Ostendorf(2018)}]{Jaech2018}
Aaron Jaech and Mari Ostendorf. 2018.
\newblock Personalized language model for query auto-completion.
\newblock In \emph{ACL}.

\bibitem[{Kang and Kim(2003)}]{kang2003query}
In-Ho Kang and GilChang Kim. 2003.
\newblock Query type classification for web document retrieval.
\newblock In \emph{SIGIR}.

\bibitem[{Kim(2014)}]{kim2014}
Yoon Kim. 2014.
\newblock Convolutional neural networks for sentence classification.
\newblock In \emph{EMNLP}.

\bibitem[{Lafferty et~al.(2001)Lafferty, McCallum, and Pereira}]{lafferty2001}
John Lafferty, Andrew McCallum, and Fernando~CN Pereira. 2001.
\newblock Conditional random fields: Probabilistic models for segmenting and
  labeling sequence data.
\newblock In \emph{ICML}.

\bibitem[{Lample et~al.(2016)Lample, Ballesteros, Subramanian, Kawakami, and
  Dyer}]{lample2016}
Guillaume Lample, Miguel Ballesteros, Sandeep Subramanian, Kazuya Kawakami, and
  Chris Dyer. 2016.
\newblock Neural architectures for named entity recognition.
\newblock \emph{NAACL}.

\bibitem[{LeCun et~al.(2015)LeCun, Bengio, and Hinton}]{lecun2015}
Yann LeCun, Yoshua Bengio, and Geoffrey Hinton. 2015.
\newblock Deep learning.
\newblock \emph{nature}.

\bibitem[{Li et~al.(2019)Li, Zhang, Bendersky, Deng, Metzler, and
  Najork}]{li2019}
Cheng Li, Mingyang Zhang, Michael Bendersky, Hongbo Deng, Donald Metzler, and
  Marc Najork. 2019.
\newblock Multi-view embedding-based synonyms for email search.
\newblock In \emph{SIGIR}.

\bibitem[{Li et~al.(2008)Li, Wang, and Acero}]{li2008learning}
Xiao Li, Ye-Yi Wang, and Alex Acero. 2008.
\newblock Learning query intent from regularized click graphs.
\newblock In \emph{SIGIR}.

\bibitem[{Mikolov et~al.(2010)Mikolov, Karafi{\'a}t, Burget,
  {\v{C}}ernock{\`y}, and Khudanpur}]{Mikolov2010}
Tom{\'a}{\v{s}} Mikolov, Martin Karafi{\'a}t, Luk{\'a}{\v{s}} Burget, Jan
  {\v{C}}ernock{\`y}, and Sanjeev Khudanpur. 2010.
\newblock Recurrent neural network based language model.
\newblock In \emph{InterSpeech}.

\bibitem[{Mitra and Craswell(2015)}]{mitra2015}
Bhaskar Mitra and Nick Craswell. 2015.
\newblock Query auto-completion for rare prefixes.
\newblock In \emph{CIKM}.

\bibitem[{Mitra et~al.(2018)Mitra, Craswell et~al.}]{mitra2018}
Bhaskar Mitra, Nick Craswell, et~al. 2018.
\newblock An introduction to neural information retrieval.
\newblock \emph{Foundations and Trends{\textregistered} in Information
  Retrieval}.

\bibitem[{Pang and Lee(2005)}]{pang2005}
Bo~Pang and Lillian Lee. 2005.
\newblock Seeing stars: Exploiting class relationships for sentiment
  categorization with respect to rating scales.
\newblock In \emph{ACL}.

\bibitem[{Park and Chiba(2017{\natexlab{a}})}]{park2017}
Dae~Hoon Park and Rikio Chiba. 2017{\natexlab{a}}.
\newblock A neural language model for query auto-completion.
\newblock In \emph{SIGIR}.

\bibitem[{Park and Chiba(2017{\natexlab{b}})}]{Park:17}
Dae~Hoon Park and Rikio Chiba. 2017{\natexlab{b}}.
\newblock A neural language model for query auto-completion.
\newblock In \emph{SIGIR}.

\bibitem[{Ramanath et~al.(2018)Ramanath, Inan, Polatkan, Hu, Guo, Ozcaglar, Wu,
  Kenthapadi, and Geyik}]{ramanath2018}
Rohan Ramanath, Hakan Inan, Gungor Polatkan, Bo~Hu, Qi~Guo, Cagri Ozcaglar,
  Xianren Wu, Krishnaram Kenthapadi, and Sahin~Cem Geyik. 2018.
\newblock Towards deep and representation learning for talent search at
  linkedin.
\newblock In \emph{CIKM}.

\bibitem[{Ren et~al.(2018)Ren, Ni, Malik, and Ke}]{Ren2018}
Gary Ren, Xiaochuan Ni, Manish Malik, and Qifa Ke. 2018.
\newblock Conversational query understanding using sequence to sequence
  modeling.
\newblock In \emph{WWW}.

\bibitem[{Sang and De~Meulder(2003)}]{sang2003}
Erik~F Sang and Fien De~Meulder. 2003.
\newblock Introduction to the conll-2003 shared task: Language-independent
  named entity recognition.
\newblock In \emph{NAACL: HLT}.

\bibitem[{Sarawagi and Cohen(2005)}]{sarawagi2005}
Sunita Sarawagi and William~W Cohen. 2005.
\newblock Semi-markov conditional random fields for information extraction.
\newblock In \emph{NeurIPS}.

\bibitem[{Sethy et~al.(2015)Sethy, Chen, Arisoy, and Ramabhadran}]{sethy2015}
Abhinav Sethy, Stanley Chen, Ebru Arisoy, and Bhuvana Ramabhadran. 2015.
\newblock Unnormalized exponential and neural network language models.
\newblock In \emph{ICASSP}.

\bibitem[{Sutskever et~al.(2014)Sutskever, Vinyals, and Le}]{Sutskever2014}
Ilya Sutskever, Oriol Vinyals, and Quoc~V. Le. 2014.
\newblock Sequence to sequence learning with neural networks.
\newblock In \emph{NeurIPS}.

\bibitem[{Wang et~al.(2018)Wang, Zhang, Mohan, Dhillon, and Kolter}]{Wang2018}
Po-Wei Wang, Huan Zhang, Vijai Mohan, Inderjit~S. Dhillon, and J.~Zico Kolter.
  2018.
\newblock Realtime query completion via deep language models.
\newblock In \emph{SIGIR Workshop On eCommerce}.

\bibitem[{Yin et~al.(2016)Yin, Hu, Tang, Daly, Zhou, Ouyang, Chen, Kang, Deng,
  Nobata et~al.}]{yin2016}
Dawei Yin, Yuening Hu, Jiliang Tang, Tim Daly, Mianwei Zhou, Hua Ouyang,
  Jianhui Chen, Changsung Kang, Hongbo Deng, Chikashi Nobata, et~al. 2016.
\newblock Ranking relevance in yahoo search.
\newblock In \emph{KDD}.

\end{thebibliography}

\end{document}